\documentclass[11pt]{article}

\usepackage[final]{acl}
\usepackage{times}
\usepackage{latexsym}

\usepackage[T1]{fontenc}

\usepackage[utf8]{inputenc}

\usepackage{microtype}

\usepackage{inconsolata}

\usepackage{graphicx}

\usepackage{amsmath}      
\usepackage{amssymb}      
\usepackage{bm}           
\usepackage{multirow}

\title{A Multimodal Text- and Graph-Based Approach for Open-Domain Event Extraction from Documents}

\author{Praval Sharma \\
  College of Information Science \& Technology\\
  University of Nebraska Omaha, USA \\
  }

\begin{document}
\maketitle
\begin{abstract}
Event extraction is essential for event understanding and analysis. It supports tasks such as document summarization and decision-making in emergency scenarios. However, existing event extraction approaches have limitations: (1) closed-domain algorithms are restricted to predefined event types and thus rarely generalize to unseen types and (2) open-domain algorithms, capable of handling unconstrained event types, have largely overlooked the potential of large language models (LLMs) despite their advanced abilities. Additionally, they do not explicitly model document-level contextual, structural, and semantic reasoning, which are crucial for effective event extraction but remain challenging for LLMs due to lost-in-the-middle phenomenon and attention dilution. To address these limitations, we propose multimodal open-domain event extraction, MODEE, a novel approach for open-domain event extraction that combines graph-based learning with text-based representation from LLMs to model document-level reasoning. Empirical evaluations on large datasets demonstrate that MODEE outperforms state-of-the-art open-domain event extraction approaches and can be generalized to closed-domain event extraction, where it outperforms existing algorithms.
\end{abstract}

\section{Introduction}

Event extraction involves identifying an event in text and extracting related arguments such as location, time, and agents involved. It is crucial for event analysis and understanding, and can be used in planning and responding to extreme phenomena such as disease outbreaks and natural disasters \cite{yuSpatiotemporalEventDetection2020}. Additionally, it is useful for tasks such as summarizing documents, clustering them, and building knowledge graphs.

There are several existing event extraction approaches. The closed-domain sentence-level event extraction approaches \cite{duEventExtractionAnswering2020, hsuDEGREEDataEfficientGenerationBased2022} use a sentence to extract events. Since events are generally described in multiple sentences \cite{tongDocEELargeScaleFinegrained2022}, they are not able to extract complete information about events. The closed-domain document-level event extraction approaches \cite{duDocumentLevelEventRole2020, liuDocumentLevelEventArgument2023} address this limitation. However, they rely on predefined event schemas tailored to specific event types and thus have limited ability to generalize to unseen types. Open-domain event extraction approaches \cite{hamborgGiveme5W1HUniversalSystem2019, liuOpenDomainEvent2019} can extract unrestricted event types and are critical for improving natural language understanding and downstream applications \cite{arakiOpenDomainEventDetection2018}. However, existing approaches primarily rely on heuristics or rule-based methods. As a result, they struggle when events are described in ways not covered by their rules. Recent advances in natural language processing, particularly in large language models (LLMs), have led to significant improvements in various information extraction tasks. Despite this, these methods remain underexplored in document-level open-domain event extraction. This task requires understanding the document-level context, structure, and semantic relationships of event-related tokens. However, LLMs often struggle to capture these aspects due to the lost-in-the-middle phenomenon and attention dilution \cite{liEventExtractionLarge2025}. Additionally, recent efforts \cite{cao5W1HExtractionLarge2024} use LLMs with simple fine-tuning, but do not explicitly model document-level reasoning or leverage multimodal learning to derive richer information for improved event extraction.

In this research, we propose multimodal open-domain event extraction (MODEE), a graph neural network (GNN)- and LLM-based approach designed for document-level event extraction in open-domain settings. MODEE combines token-level embeddings from the LLM with node-level embeddings from the GNN using an attention-based gated fusion mechanism to model document-level context, structure, and semantics of event-related tokens and extracts events in an end-to-end generative manner. This approach is motivated by recent work in retrieval augmented generation \cite{huGRAGGraphRetrievalAugmented2025} and question answering \cite{heGRetrieverRetrievalAugmentedGeneration2024}, where graph-based representations have enhanced LLM reasoning and contextual understanding. Experimental results on a large, manually annotated, and statistically verified dataset show that MODEE outperforms fine-tuned LLMs, LLMs under zero- and five-shot prompting, and an existing open-domain event extraction approach. Additional experiments on a large closed-domain document-level event extraction dataset demonstrate the generality and adaptability of MODEE in closed-domain settings, although it is primarily designed for open-domain settings. The main contributions of this research are: (1) a novel graph neural network- and LLM-based approach, MODEE, for open-domain document-level event extraction and (2) an attention-based gated multimodal fusion mechanism that integrates token-level LLM embeddings with node-level graph embeddings for effective event extraction.

Note that, MODEE extracts the five key elements, i.e., 5Ws (where, when, what, who, and why), of events from documents. It is designed to extract the 5Ws because the 5Ws framework forms the foundation of event reporting in documents, particularly in news media \cite{harrowerReportingPracticalGuide2010}, applies to unconstrained event types, and is used by prior open-domain event extraction approaches  \cite{hamborgGiveme5WMainEvent2018, liuOpenDomainEvent2019}. MODEE follows the \textit{one-event-per-document} setting and extracts the main event as in \citet{tongDocEELargeScaleFinegrained2022}. While documents contain multiple events, they typically center on a newsworthy main event, with background events providing supporting information. Therefore, extracting the main event is critical for news discourse comprehension and has been used as the core unit of analysis in news discourse studies \cite{choubeyDiscourseFunctionEvent2020} and event-centric clustering \cite{zhangEnhancingEventcentricNews2025}.

\section{Related Work}
\textbf{Closed-Domain Event Extraction: }These approaches rely on predefined event schemas to extract events from text. For example, they use the schema for a ‘Conflict-Attack’ event that defines arguments such as ‘attacker,’ ‘target,’ and ‘instrument’ to extract this type of event. The closed-domain sentence-level event extraction approaches rely on hand-crafted features \cite{liJointEventExtraction2013}, convolutional neural networks \cite{chenEventExtractionDynamic2015}, recurrent neural networks \cite{liuExploitingArgumentInformation2017}, graph-based techniques \cite{nguyenJointExtractionEntities2022}, question-answering \cite{duEventExtractionAnswering2020}, and generative modeling \cite{caoZeroShotCrossLingualEvent2023}. Because events are often described across multiple sentences \cite{tongDocEELargeScaleFinegrained2022}, these sentence-level methods often fail to extract complete event information. To address this, document-level approaches relying on machine reading comprehension \cite{duDocumentLevelEventRole2020}, deep valued networks \cite{huangDocumentlevelEventExtraction2021}, chain reasoning \cite{liuDocumentLevelEventArgument2023}, and graph-based methods \cite{huangIterativelyParallelGeneration2023, wanJointDocumentLevelEvent2023} have been proposed. While these operate at the document level, they typically rely on predefined event schemas and thus struggle to generalize to events outside the schema set. Graph-based methods further rely on multi-step processes, where entities are first identified, followed by graph construction and event extraction. In contrast, our approach employs the 5Ws framework, which enables it to extract unrestricted event types, and performs end-to-end event extraction by integrating graph representations into a generative model. It constructs a document-level token graph without relying on external entity identification and jointly models graph and textual information for multimodal reasoning. This end-to-end generative design distinguishes it from prior graph-based methods.

\textbf{Open-Domain Event Extraction: }These approaches extract events without relying on a predefined set of event schemas. They are therefore able to extract unconstrained types of events. They employ various techniques, including heuristics and linguistic rules \cite{hamborgGiveme5W1HUniversalSystem2019}, distant supervision \cite{arakiOpenDomainEventDetection2018}, clustering \cite{huangLiberalEventExtraction2016}, Bayesian models \cite{yuanOpenSchemaEventProfiling2018}, neural latent variable models \cite{liuOpenDomainEvent2019}, adversarial domain adaptation \cite{naikOpenDomainEvent2020}, and bi-directional LSTM \cite{veysehAugmentingOpenDomainEvent2021}. They rely on handcrafted rules or limited supervision and thus face generalization challenges across diverse content and writing styles. Various information extraction tasks have benefitted from the use of LLMs. However, their application to open-domain event extraction remains limited. LLMs often struggle to capture document-level context, structure, and semantics of event-related tokens because of the lost-in-the-middle phenomenon and attention dilution, which are critical for event extraction \cite{liEventExtractionLarge2025}. However, recent studies \cite{cao5W1HExtractionLarge2024} that apply LLMs use simple fine-tuning and do not explicitly model these document-level aspects. In contrast, our approach integrates graph-based learning with text-based representation from LLMs to model these aspects for improved event extraction.

\textbf{Multimodal Event Extraction: }These approaches leverage multiple modalities, such as images and text, to extract events. They employ techniques such as vision-language models \cite{liCLIPEventConnectingText2022}, attention-based fusion \cite{sunUMIEUnifiedMultimodal2024}, and modality-shared encoder \cite{caoCrossmodalMultitaskLearning2025}. While using multiple modalities, particularly images and text, has proven effective, their utilization requires parallel annotations across modalities. This is both expensive and complex \cite{caoCrossmodalMultitaskLearning2025}, and therefore difficult to scale. As a result, most existing multimodal approaches are generally trained on image-text pairs without parallel annotations and confined to closed-domain event extraction. In contrast, in this study, we explore a more scalable and generalizable approach that derives complementary modalities from a single annotated source and uses them for open-domain event extraction. Specifically, we construct a graph using a document’s text and use the resulting text-graph pairs for event extraction.

\section{Methodology}
\subsection{Problem Definition}
Given a document $D$, which describes an event $e$, the goal of open-domain document-level event extraction is to extract five key elements of $e$, i.e., $5Ws_e$ ($where_e$, $when_e$, $what_e$, $who_e$, $why_e$), based on the information in $D$.

\subsection{Multimodal Open-Domain Event Extraction}
\subsubsection{Overview of the Approach}
As shown in Figure \ref{fig:figure1}, multimodal open-domain event extraction (MODEE) consists of four modules: (1) Text encoder that produces contextual token-level text embeddings for a document, (2) Graph encoder that produces node-level graph embeddings for a document-level token graph created using a document, (3) Attention-based gated multimodal fusion module that integrates text and graph embeddings to produce integrated multimodal embeddings, and (4) Text decoder that generates 5Ws for the main event described in a document conditioned on the integrated multimodal embeddings. They are described in the following sections.
\subsubsection{Text Encoder}
The text encoder in MODEE encodes the token sequence from a given document to contextualized embeddings. Given a document $D$ with $n$ tokens $\{t_1, t_2, \ldots, t_n\}$, the encoder produces token-level hidden representations $\boldsymbol{H}_{\textit{text}} \in \mathbb{R}^{n \times d}$, where $d$ is the hidden dimension:
\begin{equation}
    \boldsymbol{H}_{\textit{text}} = \mathit{TextEncoder}(D).
    \label{eq:1}
\end{equation}

In MODEE, we use the encoder part of T5 \cite{raffelExploringLimitsTransfer2020} as the text encoder and the decoder part as the text decoder (see Section 3.2.5). T5's encoder-decoder architecture allows integration of representations from multiple modalities (e.g., text and graph) and  generation of task-specific outputs (e.g., event 5Ws) conditioned on the integrated representations. This makes it particularly well-suited for integrating multimodal data and generative inference for event extraction, which is the focus of this study. While we use T5, other encoder-decoder LLMs such as Flan-T5 \cite{chungScalingInstructionFinetunedLanguage2024} can also be incorporated into MODEE.

\subsubsection{Graph Encoder}
The graph encoder in MODEE produces node-level embeddings for a document-level token graph created using a document. Given a document $D = \{t_1, t_2, \ldots, t_n\}$ with $n$ tokens, it first creates a document-level token graph $G=(V, E)$, where $V$ is a set of vertices representing the tokens in $D$, \begin{figure}[t]
  \includegraphics[width=\columnwidth]{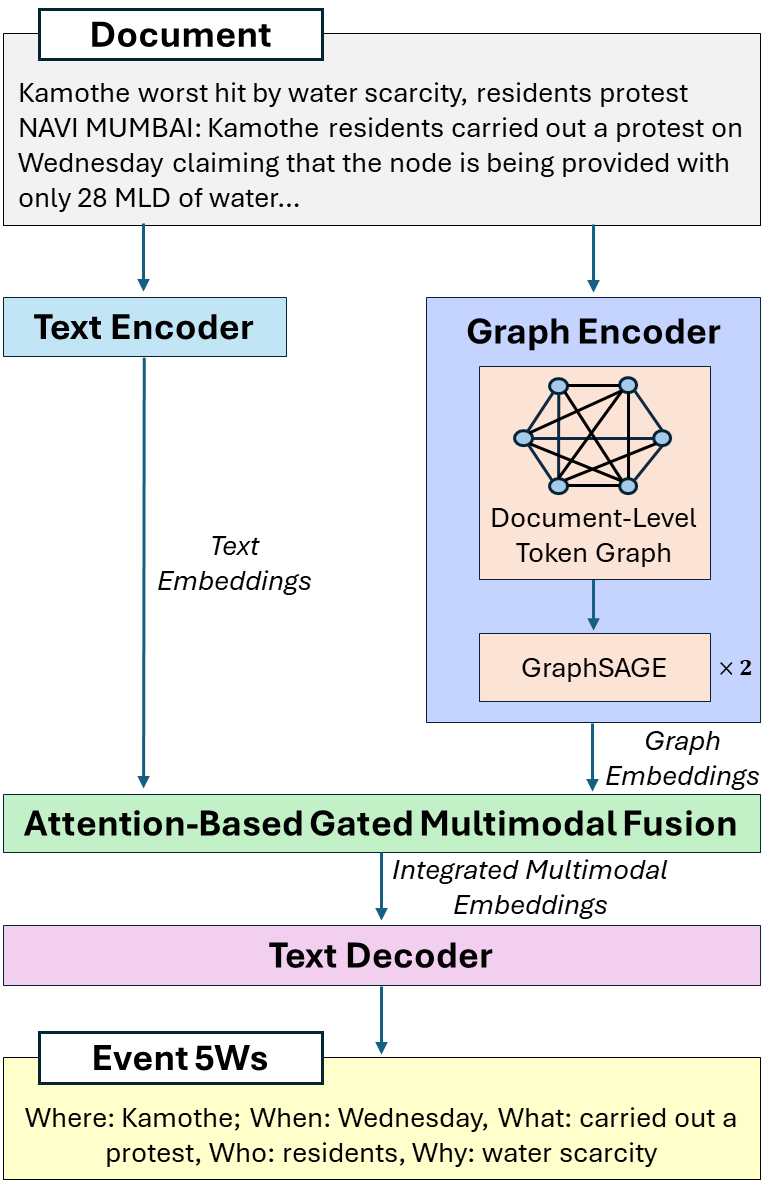}
  \caption{Overview of event extraction in MODEE.}
  \label{fig:figure1}
\end{figure} i.e., $V=\{t_1,t_2, \ldots,t_n\}$, and $E$ is the set of undirected edges between every pair of vertices, i.e., $E = \{ (t_i, t_j) \mid t_i, t_j \in V,\; i, j = 1, 2, \ldots, n,\; i \neq j \}$, forming a complete graph. Then, it produces embeddings, $\boldsymbol{H}_{\textit{graph}} \in \mathbb{R}^{n \times d}$, for the graph using a two-layer GraphSAGE \cite{hamiltonInductiveRepresentationLearning2017} with LSTM aggregation, where $n$ is the number of nodes in the graph, which is also the total number of tokens in $D$, and $d$ is the hidden dimension:
\begin{equation}
    \boldsymbol{H}_{\textit{graph}} = \mathit{GraphEncoder}(D).
    \label{eq:2}
\end{equation}

The complete graph allows the graph encoder to model long-range dependencies between tokens across an entire document. By using GraphSAGE to encode the graph, the encoder then effectively captures the document-level structure and semantic relationships of event-related tokens in latent space. This helps MODEE determine the relevance of tokens in a document for event extraction.

\subsubsection{Attention-Based Gated Multimodal Fusion}
This module in MODEE integrates token-level contextual embeddings from the text encoder with document-level structure- and semantic-aware node-level representations from the graph encoder to produce rich integrated multimodal embeddings. The integration process involves two steps: Attention-based gating vector computation and Integrated multimodal embedding computation.

\textbf{\textit{Attention-Based Gating Vector Computation:}} In this step, MODEE computes attention scores and generates a gating vector based on the text and graph embeddings to determine the relevance of individual tokens in a document with respect to the event described. This process is inspired by additive attention \cite{bahdanauNeuralMachineTranslation2015}, which facilitates richer interaction between representations from multiple modalities. Given $\boldsymbol{H}_{\textit{text}} \in \mathbb{R}^{n \times d}$, the token embeddings from the text encoder (Equation \ref{eq:1}), and $\boldsymbol{H}_{\textit{graph}} \in \mathbb{R}^{n \times d}$, the node embeddings from the graph encoder (Equation \ref{eq:2}), MODEE first projects these embeddings into a shared latent space using linear layers to enable direct cross-modal interaction: 
\begin{equation}
\boldsymbol{H}_{\textit{textProj}} = \boldsymbol{W}_{\textit{textProj}} \boldsymbol{H}_{\textit{text}}
\label{eq:3}
\end{equation}
\begin{equation}
\boldsymbol{H}_{\textit{graphProj}} = \boldsymbol{W}_{\textit{graphProj}} \boldsymbol{H}_{\textit{graph}}
\label{eq:4}
\end{equation}
where $\boldsymbol{W}_{\textit{textProj}}$, $\boldsymbol{W}_{\textit{graphProj}}$  $\in \mathbb{R}^{d \times d}$  are the weights of the linear layers. Although both text and graph embeddings have the same dimension in our setup, this projection supports modality fusion in settings where the embedding dimensions differ and enhances the generalizability of MODEE.

Next, it computes hidden representation $\boldsymbol{H}_{\textit{hidden}}$ through an element-wise addition of $\boldsymbol{H}_{\textit{textProj}}$ and $\boldsymbol{H}_{\textit{graphProj}}$ followed by a non-linear activation:
\begin{equation}
\boldsymbol{H}_{\textit{hidden}} = \tanh \Big( \boldsymbol{H}_{\textit{textProj}} + \boldsymbol{H}_{\textit{graphProj}} \Big).
\label{eq:5}
\end{equation}
This hidden representation captures both the contextual information from the text and the structural and semantic cues from the graph. Finally, to compute the token-wise gating vector $\alpha \in \mathbb{R}^{n \times 1}$, MODEE projects $\boldsymbol{H}_{\textit{hidden}}$ to a vector of scalar scores using a linear layer, $\boldsymbol{V}_{\textit{attn}} \in \mathbb{R}^{d \times 1}$ followed by a sigmoid activation:
\begin{equation}
\boldsymbol{\alpha} = \sigma \big( \boldsymbol{H}_{\textit{hidden}} \cdot \boldsymbol{V}_{\textit{attn}} \big)
\label{eq:6}
\end{equation}
where $\sigma(\cdot)$ denotes the sigmoid function. Since $\boldsymbol{H}_{\textit{hidden}}$ is used to compute $\boldsymbol{\alpha}$, the cues from both text and graph embeddings are reflected in the gating scores and higher scores are assigned for tokens most relevant to the event described in $D$.

\textbf{Integrated Multimodal Embedding Computation:} In this step, MODEE produces integrated multimodal embeddings $\boldsymbol{H}_{\textit{integrated}} \in \mathbb{R}^{n \times d}$ by applying the gating vector $\boldsymbol{\alpha}$ to the original text embeddings $\boldsymbol{H}_{\textit{text}}$:
\begin{equation}
\boldsymbol{H}_{\textit{integrated}} = \boldsymbol{H}_{\textit{text}} \odot \boldsymbol{\alpha}
\label{eq:7}
\end{equation}
where $\odot$ denotes element-wise multiplication. This helps highlight the event-related tokens in a document. As a result, $\boldsymbol{H}_{\textit{integrated}}$, when passed to the decoder, enables MODEE to selectively focus on tokens with high relevance to the event described in $D$ and supports more accurate event extraction.

\subsubsection{Text Decoder}
The text decoder in MODEE generates 5Ws for the event described in a document. Given $\boldsymbol{H}_{\textit{integrated}} \in \mathbb{R}^{n \times d}$, the integrated multimodal embeddings produced by the attention-based gated fusion module for a document $D$ describing an event, the text decoder autoregressively generates the 5Ws, i.e., where, when, what, who, and why: 
\begin{equation}
\text{5Ws} = \mathit{TextDecoder}(\boldsymbol{H}_{\textit{integrated}})
\label{eq:8}
\end{equation}

In this work, we use the decoder part of the same T5 architecture used as the text encoder (see Section 3.2.2) to form a unified encoder-decoder framework. At inference time, the decoder generates 5Ws in a sequence-to-sequence manner beginning with a start token \texttt{<pad>} as initial decoder input and terminating when the end-of-sequence token \texttt{</s>} is generated. The decoder’s output format is “where:\texttt{<>}; when:\texttt{<>}; what:\texttt{<>}; who:\texttt{<>}; why:\texttt{<>},” where each placeholder \texttt{<>} is replaced with the text generated for the corresponding 5W class or with ‘none’ if no text is generated, as shown in Figure \ref{fig:figure1}.

\subsection{Training Strategy}
We train the four modules of MODEE jointly in an end-to-end fashion so that they support the unified objective of accurate event extraction. The text encoder, the attention-based gated multimodal fusion module, and the text decoder are optimized using cross-entropy loss:
\begin{equation}
\mathcal{L}_{\text{crossentropy}} = - \sum_{t=1}^{m} 
\log P\big(y_t \mid y_t^*, \boldsymbol{H}_{\textit{integrated}}\big)
\label{eq:9}
\end{equation}
where $m$ is the number of ground-truth tokens, $y_t$ is the ground-truth token at position $t$, $y_t^*$ denotes the sequence of all previous tokens, i.e.,
$(y_{t-1}, y_{t-2}, \dots, y_2, y_1)$, and $P(\cdot)$ is the probability assigned by the model to $y_t$ conditioned on the integrated multimodal embeddings $\boldsymbol{H}_{\textit{integrated}}$. This training design allows the encoder to learn informative contextual representations, the fusion module to generate rich integrated multimodal embeddings conditioned on both text- and graph-based cues, and the decoder to generate accurate event 5Ws.

The graph encoder is optimized using contrastive loss \cite{chenSimpleFrameworkContrastive2020}. Given $\boldsymbol{H}_{\textit{graph}} = \{\boldsymbol{z}_1, \boldsymbol{z}_2, \dots, \boldsymbol{z}_n\}$, the embeddings for $n$ nodes $\{t_1, t_2, \dots, t_n\}$ from a document-level token graph $G$, and $y = \{y_1, y_2, \dots, y_n\}$, their corresponding 5W classes, the contrastive loss for each node pair $(t_i, t_j)$ such that $y_i = y_j$, $i \neq j$, and $1 \leq i \leq j \leq n$ is computed as follows:
\begin{equation}
\mathcal{L}_{\text{contrastive}} = - \log \frac{\exp\big(\textit{sim}(\boldsymbol{z}_i, \boldsymbol{z}_j) / \tau \big)}{\sum_{l=1}^{n} \exp\big(\textit{sim}(\boldsymbol{z}_i, \boldsymbol{z}_l) / \tau \big)}
\label{eq:10}
\end{equation}
where $\boldsymbol{z}_i$ and $\boldsymbol{z}_j$ are the embeddings for nodes $t_i$ and $t_j$, respectively, $\boldsymbol{z}_l$ are the embeddings for nodes such that $y_l \neq y_i$, $\textit{sim}(\cdot)$ denotes cosine similarity, and $\tau$ is a temperature hyperparameter to adjust the loss function’s sensitivity. This training strategy allows the graph encoder to learn to draw together tokens representing the same 5W class (e.g., all “where” tokens) and push apart those from different classes in the embedding space. As a result, the encoder learns to automatically capture the document-level structure and semantics of event-related tokens in latent space and provide valuable cues to the attention-based gated multimodal fusion module for computing the gating vector and determining the relevance of each token in a document with respect to the event described. Due to memory constraints, we compute contrastive learning on a sampled subset of nodes during training. Specifically, we randomly sample five nodes for each of the 5W classes and an additional set of five nodes that do not belong to any 5W class. This sampling strategy balances efficiency with class diversity and enables the graph encoder to learn discriminative representations. Note that we initialize each node in the document-level token graph using the encoder of the pretrained T5 model during training.

\section{Experiments}
\subsection{Dataset and Evaluation}
We developed a dataset to train and evaluate MODEE. It consists of 10,000 news reports published between 2015 to 2019 from seven Indian newspapers: Times of India, The Hindu, The Pioneer, Economic Times, Assam Tribune, Kashmir Observer, and Incredible Orissa. These newspapers vary in journalistic style and geographic focus, and thus contribute to the diversity of the dataset.

According to the inverse pyramid concept, key information typically appears early in a report \cite{harrowerReportingPracticalGuide2010}. Additionally, \citet{ebnerMultiSentenceArgumentLinking2020} observed that majority of the arguments for an event appeared within a five-sentence window in reports. Guided by these observations and because we focus on extracting the main event, we included the title and first five sentences of each report in our dataset. This allowed us to balance annotation cost while including the portions of reports where main events are most frequently described.

The dataset was created in three stages: training, annotation, and resolution. In the training stage, three coders (university students familiar with Indian context) underwent three rounds of training to familiarize them with the task. In each round, they annotated seven reports (one from each of the seven newspapers) not included in the final dataset. After each round, they participated in discussions to resolve differences in their annotations and improve consistency. By the final round, they achieved an inter-coder reliability above 0.8, measured using Krippendorff’s alpha \cite{krippendorffReliabilityMultiValuedCoding2016}, which indicates high consistency.

In the annotation stage, the coders independently annotated all 10,000 reports in the dataset. In the resolution stage, they resolved annotation differences based on a defined policy. If all coders agreed on an annotation, it was considered the gold standard. If two coders agreed on an annotation, the third was asked to reannotate. If unanimous agreement was subsequently reached, the annotation was considered the gold standard. If disagreement persisted, an expert (a researcher familiar with event extraction) determined the gold standard. If all coders disagreed on an annotation, all reannotated, and the process was repeated until at least two coders reached an agreement.

Once the dataset development was completed, we randomly split it into training, validation, and test sets containing 8,000, 1,000, and 1,000 reports, respectively (see Appendix \ref{appen1} for dataset details). The training and validation sets were used to train MODEE and the test set was used for evaluation. In this research, we report the evaluation results using precision ($P$), recall ($R$), and $F1$ score (see Section 4.4) computed using exact match (EM) following prior work on event extraction \cite{liuOpenDomainEvent2019, tongDocEELargeScaleFinegrained2022}. We also report results using ROUGE-L \cite{linROUGEPackageAutomatic2004} and BERTScore \cite{zhangBERTScoreEvaluatingText2020} to capture lexical and semantic similarity between predictions and gold standard, with precision, recall, and F1 score computed per document and averaged over the test set.

\subsection{Training Configuration}
The text encoder and decoder modules are trained with AdamW optimizer using learning rate of 1e-3 (T5-Small) and 1e-4 (T5-Base). Input and output sequence lengths are capped at 512 tokens and beam size is set to 5 during inference. The graph encoder and attention-based gated fusion modules are trained using Adam optimizer with 1e-3 learning rate and 5e-4 weight decay. MODEE is trained for 10 epochs on NVIDIA V100 GPUs, using gradient accumulation to achieve an effective batch size of 8 due to memory constraints.

\subsection{Baselines}
We evaluate the efficacy of MODEE by comparing it against several baselines. Although MODEE uses multiple modalities (i.e., graph and text), both are derived directly from textual data unlike existing multimodal approaches that rely on multiple sources (e.g., image-text pairs). To ensure a fair comparison, we therefore evaluate MODEE against baselines that operate solely on document-level text. As existing LLM-based open-domain event extraction approaches are not publicly available, we fine-tune standard T5 models  \cite{raffelExploringLimitsTransfer2020}, T5-Small, T5-Base, and T5-Large, on our task using the configuration described in Section 4.2 (with T5-Large trained using the same configuration as T5-Base). We also compare MODEE against modern strong pretrained LLMs such as Llama 3.1 (8B and 70B), Qwen 3 (8B and 32B), and Mistral V0.3 (7B) under zero- and five-shot prompting\footnote{We use 8-bit quantization to reduce GPU memory usage.} (see Appendix \ref{appen2} for prompts used). For a fair comparison, we further fine-tune smaller LLMs, Llama 3.2 (1B) and Qwen 3 (0.6B), which are comparable in size to our approach, using the same configuration as T5-Base (Section 4.2). In addition, we use Giveme5W1H \cite{hamborgGiveme5W1HUniversalSystem2019}, a heuristic-based open-source open-domain event extraction algorithm, as a non-generative baseline.

\subsection{Main Result}
Table \ref{tab:table1} presents the performance of MODEE and all baselines on the test dataset. MODEE-Base, which uses T5-Base, achieves the best overall performance across EM, ROUGE-L, and BERTScore. Despite being based on the same architecture, its improved performance over T5-Base demonstrates that incorporating multiple modalities derived from single source, particularly creating graphs from text and using the text-graph pairs, enhances open-domain event extraction. The better performance of MODEE-Small compared to T5-Small further bolsters this. T5-Large, despite being a larger model, does not outperform MODEE-Base. This shows that using a larger model does not compensate for the lack of document-level context, structure, and semantics of event-related tokens that MODEE captures through multimodal integration. Giveme5W1H is the least effective model and illustrates the limitations of rule-based approaches.

Modern LLMs do not perform well under both zero- and five-shot prompting. This indicates that prompting alone is insufficient for accurately extracting events. Increasing model size does not lead to improved performance. Additionally, fine-tuned LLMs, Llama 1B and Qwen 0.6B, underperform compared to MODEE, despite being trained under the same configuration and having more parameters. This suggests that effective event extraction requires capturing document-level context, structure, and semantics of event-related tokens, which MODEE does by combining text- and graph-based representations. Its higher scores on ROUGE-L and BERTScore further indicate that it produces more semantically coherent 5Ws than the LLMs.

To provide a more fine-grained analysis, we examine model performance across individual 5W classes (see Appendix \ref{appen3}). While some models perform better on specific Ws (e.g., T5-Large on When), MODEE consistently achieves stronger performance on Why, which exhibits higher lexical variability, greater sparsity, and longer spans (see Appendix \ref{appen1}), and therefore requires deeper understanding of document-level context, structure, and semantics. MODEE leverages contrastive learning in its graph encoder to pull tokens with similar semantic roles closer in the embedding space, and integrates structural graph representations from the document-level token graph with contextual textual representations to capture  document-level context, structure, and semantics of event-related tokens. This allows MODEE to outperform all baselines even on the more challenging Why. 

MODEE gains from scaling the underlying language model. Transitioning from T5-Small to T5-Base improved its performance. This shows the versatility and scalability of MODEE in effectively leveraging more powerful language models.

\subsection{Ablation Study}
To investigate the contributions of different modules in MODEE, we created ablated versions of MODEE-Base, our best performing model. First, to evaluate the role of contrastive learning in enhancing the graph encoder and the downstream event extraction, we train MODEE \textbf{without contrastive learning}. In this setting, the graph encoder is trained jointly with all other modules using the same cross-entropy loss (see Equation \ref{eq:9}). Next, to assess the importance of attention-based gated multimodal fusion, we replace the fusion module with simple \textbf{element-wise addition} of the text and graph embeddings produced by the respective encoders. Finally, to examine the effect of document-level structural information derived from the complete graph, we modify the graph encoder to operate on a \textbf{linear graph}, where each token is connected only to its immediate neighbors in the document forming a linear chain. In this setting, the graph encoder can only capture local token context (i.e., context within a small window), which limits its ability to capture document-level event structure.

Table \ref{tab:table2} presents the results of the ablation study. Removing contrastive learning leads to performance drops across all 5W classes compared to the full MODEE. This demonstrates that contrastive learning helps the graph encoder produce more meaningful embeddings by pulling together semantically similar tokens in the embedding space. As a result, MODEE can more effectively focus on relevant event-related tokens during integration, thereby leading to a more accurate event extraction. Replacing the attention-based gated multimodal fusion module with element-wise addition results in a significant decline in performance. This highlights the importance of attention-based gating mechanism in integrating text and graph representations. Additionally, using a linear graph instead of the document-level complete graph leads to a substantial performance drop. This illustrates the importance of the document-level structural context for accurate event extraction. Overall, the full MODEE outperforms all ablated models, which validates the efficacy of its different modules.

\begin{table*}[t]
\centering
\small
\begin{tabular}{l|lll|lll|lll}
\hline
\multicolumn{1}{c|}{\multirow{2}{*}{\textbf{Models}}} & \multicolumn{3}{c|}{\textbf{Exact Match}} & \multicolumn{3}{c|}{\textbf{ROUGE-L}} & \multicolumn{3}{c}{\textbf{BERTScore}} \\ \cline{2-10} 
\multicolumn{1}{c|}{} & \multicolumn{1}{l|}{\textit{P}} & \multicolumn{1}{l|}{\textit{R}} & \textit{F1} & \multicolumn{1}{l|}{\textit{P}} & \multicolumn{1}{l|}{\textit{R}} & \textit{F1} & \multicolumn{1}{l|}{\textit{P}} & \multicolumn{1}{l|}{\textit{R}} & \textit{F1} \\ \hline
\textbf{T5-Small (Fine-tuned)} & \multicolumn{1}{l|}{52.2} & \multicolumn{1}{l|}{47.5} & 49.8 & \multicolumn{1}{l|}{69.8} & \multicolumn{1}{l|}{69.7} & 67.8 & \multicolumn{1}{l|}{93.5} & \multicolumn{1}{l|}{93.6} & 93.5 \\ 
\textbf{T5-Base (Fine-tuned)} & \multicolumn{1}{l|}{54.9} & \multicolumn{1}{l|}{51.8} & 53.3 & \multicolumn{1}{l|}{71.7} & \multicolumn{1}{l|}{72.8} & 70.5 & \multicolumn{1}{l|}{94.1} & \multicolumn{1}{l|}{94.2} & 94.1 \\ 
\textbf{T5-Large (Fine-tuned)} & \multicolumn{1}{l|}{57.2} & \multicolumn{1}{l|}{54.4} & 55.8 & \multicolumn{1}{l|}{\textbf{75.2}} & \multicolumn{1}{l|}{75.1} & 73.0 & \multicolumn{1}{l|}{94.5} & \multicolumn{1}{l|}{94.5} & 94.5 \\ 
\textbf{Giveme5W1H} & \multicolumn{1}{l|}{15.6} & \multicolumn{1}{l|}{16.6} & 16.1 & \multicolumn{1}{l|}{31.1} & \multicolumn{1}{l|}{31.4} & 28.8 & \multicolumn{1}{l|}{86.6} & \multicolumn{1}{l|}{86.3} & 86.3 \\ 
\textbf{Llama 8B (0-shot)} & \multicolumn{1}{l|}{9.8} & \multicolumn{1}{l|}{10.6} & 10.2 & \multicolumn{1}{l|}{27.7} & \multicolumn{1}{l|}{52.7} & 30.8 & \multicolumn{1}{l|}{84.1} & \multicolumn{1}{l|}{87.9} & 85.8 \\ 
\textbf{Llama 8B (5-shot)} & \multicolumn{1}{l|}{9.6} & \multicolumn{1}{l|}{10.3} & 10.0 & \multicolumn{1}{l|}{27.1} & \multicolumn{1}{l|}{51.8} & 30.1 & \multicolumn{1}{l|}{83.9} & \multicolumn{1}{l|}{87.8} & 85.7 \\ 
\textbf{Llama 70B (0-shot)} & \multicolumn{1}{l|}{14.1} & \multicolumn{1}{l|}{13.4} & 13.8 & \multicolumn{1}{l|}{31.2} & \multicolumn{1}{l|}{45.2} & 31.4 & \multicolumn{1}{l|}{85.2} & \multicolumn{1}{l|}{87.5} & 86.2 \\ 
\textbf{Llama 70B (5-shot)} & \multicolumn{1}{l|}{14.4} & \multicolumn{1}{l|}{13.7} & 14.0 & \multicolumn{1}{l|}{31.3} & \multicolumn{1}{l|}{45.2} & 31.5 & \multicolumn{1}{l|}{85.3} & \multicolumn{1}{l|}{87.6} & 86.3 \\ 
\textbf{Qwen 8B (0-shot)} & \multicolumn{1}{l|}{15.8} & \multicolumn{1}{l|}{19.2} & 17.3 & \multicolumn{1}{l|}{37.1} & \multicolumn{1}{l|}{55.9} & 40.5 & \multicolumn{1}{l|}{86.7} & \multicolumn{1}{l|}{89.4} & 88.0 \\ 
\textbf{Qwen 8B (5-shot)} & \multicolumn{1}{l|}{15.8} & \multicolumn{1}{l|}{19.3} & 17.4 & \multicolumn{1}{l|}{37.2} & \multicolumn{1}{l|}{56.4} & 40.5 & \multicolumn{1}{l|}{86.7} & \multicolumn{1}{l|}{89.4} & 88.0 \\ 
\textbf{Qwen 32B (0-shot)} & \multicolumn{1}{l|}{9.9} & \multicolumn{1}{l|}{11.6} & 10.7 & \multicolumn{1}{l|}{30.3} & \multicolumn{1}{l|}{68.0} & 37.2 & \multicolumn{1}{l|}{84.6} & \multicolumn{1}{l|}{89.9} & 87.1 \\ 
\textbf{Qwen 32B (5-shot)} & \multicolumn{1}{l|}{9.0} & \multicolumn{1}{l|}{10.5} & 9.7 & \multicolumn{1}{l|}{29.4} & \multicolumn{1}{l|}{67.7} & 36.5 & \multicolumn{1}{l|}{84.5} & \multicolumn{1}{l|}{89.8} & 87.0 \\ 
\textbf{Mistral 7B (0-shot)} & \multicolumn{1}{l|}{15.5} & \multicolumn{1}{l|}{18.8} & 17.0 & \multicolumn{1}{l|}{36.9} & \multicolumn{1}{l|}{51.4} & 38.6 & \multicolumn{1}{l|}{86.8} & \multicolumn{1}{l|}{89.2} & 87.9 \\ 
\textbf{Mistral 7B (5-shot)} & \multicolumn{1}{l|}{15.4} & \multicolumn{1}{l|}{18.6} & 16.8 & \multicolumn{1}{l|}{36.6} & \multicolumn{1}{l|}{51.2} & 38.5 & \multicolumn{1}{l|}{86.8} & \multicolumn{1}{l|}{89.2} & 87.9 \\ 
\textbf{Llama 1B (Fine-tuned)} & \multicolumn{1}{l|}{58.2} & \multicolumn{1}{l|}{55.3} & 56.8 & \multicolumn{1}{l|}{73.6} & \multicolumn{1}{l|}{72.5} & 71.3 & \multicolumn{1}{l|}{94.4} & \multicolumn{1}{l|}{94.4} & 94.4 \\ 
\textbf{Qwen 0.6B (Fine-tuned)} & \multicolumn{1}{l|}{56.0} & \multicolumn{1}{l|}{\textbf{57.7}} & 56.9 & \multicolumn{1}{l|}{72.8} & \multicolumn{1}{l|}{73.2} & 71.1 & \multicolumn{1}{l|}{94.3} & \multicolumn{1}{l|}{94.4} & 94.3 \\ 
\textbf{MODEE-Small} & \multicolumn{1}{l|}{54.7} & \multicolumn{1}{l|}{51.7} & 53.2 & \multicolumn{1}{l|}{71.2} & \multicolumn{1}{l|}{72.8} & 70.1 & \multicolumn{1}{l|}{93.9} & \multicolumn{1}{l|}{94.2} & 94.0 \\ 
\textbf{MODEE-Base} & \multicolumn{1}{l|}{\textbf{58.7}} & \multicolumn{1}{l|}{56.7} & \textbf{57.7} & \multicolumn{1}{l|}{75.1} & \multicolumn{1}{l|}{\textbf{75.9}} & \textbf{73.7} & \multicolumn{1}{l|}{\textbf{94.7}} & \multicolumn{1}{l|}{\textbf{94.8}} & \textbf{94.7} \\ \hline
\end{tabular}
\caption{Performance comparison on event extraction (\%) using Exact Match, ROUGE-L, and BERTScore. Bold indicates the best performance.}
\label{tab:table1}
\end{table*}

\begin{table*}[]
\centering
\resizebox{\textwidth}{!}{
\begin{tabular}{l|ccc|ccc|ccc|ccc|ccc}
\hline
\multicolumn{1}{c|}{\multirow{2}{*}{\textbf{Method}}} & \multicolumn{3}{c|}{\textbf{Where}} & \multicolumn{3}{c|}{\textbf{When}} & \multicolumn{3}{c|}{\textbf{What}} & \multicolumn{3}{c|}{\textbf{Who}} & \multicolumn{3}{c}{\textbf{Why}} \\
\multicolumn{1}{c|}{} & \textit{P} & \textit{R} & \textit{F1} & \textit{P} & \textit{R} & \textit{F1} & \textit{P} & \textit{R} & \textit{F1} & \textit{P} & \textit{R} & \textit{F1} & \textit{P} & \textit{R} & \textit{F1} \\ \hline
\textbf{MODEE} & \textbf{67.0} & \textbf{68.0} & \textbf{67.5} & \textbf{85.7} & \textbf{79.4} & \textbf{82.4} & \textbf{35.7} & \textbf{35.7} & \textbf{35.7} & \textbf{59.1} & \textbf{56.9} & \textbf{58.0} & \textbf{35.7} & \textbf{31.5} & \textbf{33.5} \\
\textbf{Without Contrastive Learning} & 62.0 & 63.7 & 62.8 & 85.0 & 79.2 & 82.0 & 34.2 & 34.2 & 34.2 & 53.9 & 52.5 & 53.2 & 31.2 & 26.4 & 28.6 \\
\textbf{Element-Wise Addition} & 7.1 & 3.5 & 4.7 & 13.8 & 14.5 & 14.1 & 10.4 & 10.4 & 10.4 & 5.6 & 5.6 & 5.6 & 2.3 & 0.2 & 0.4 \\
\textbf{Linear Graph} & 44.2 & 45.3 & 44.7 & 71.5 & 71.1 & 71.3 & 16.3 & 16.3 & 16.3 & 33.7 & 30.8 & 32.2 & 7.8 & 4.6 & 5.8 \\ \hline
\end{tabular}
}
\caption {Ablation results of MODEE. Bold indicates the best performance.}
\label{tab:table2}
\end{table*}

\begin{table}[t]
\centering
\resizebox{\columnwidth}{!}{
\begin{tabular}{l|ccc|ccc}
\hline
\multicolumn{1}{c|}{\multirow{2}{*}{\textbf{Method}}} & \multicolumn{3}{c|}{\textbf{EM}} & \multicolumn{3}{c}{\textbf{HM}} \\
\multicolumn{1}{c|}{} & \textit{P} & \textit{R} & \textit{F1} & \textit{P} & \textit{R} & \textit{F1} \\ \hline
\textbf{BERT\_Seq} & 35.3 & 35.9 & 35.6 & 20.3 & 25.0 & 22.4 \\
\textbf{MG-Reader} & 30.3 & 35.9 & 32.9 & 45.6 & 50.8 & 48.1 \\
\textbf{Doc2EDAG} & 37.1 & \textbf{36.1} & 36.6 & 54.2 & 53.7 & 53.9 \\
\textbf{BERT\_QA} & 41.9 & 28.1 & 33.5 & 75.8 & 50.6 & 60.7 \\
\textbf{Ontology\_QA} & 51.3 & 34.2 & 41.0 & 80.3 & 53.6 & 64.3 \\
\textbf{MODEE-Base} & \textbf{56.4} & 35.0 & \textbf{43.1} & \textbf{88.9} & \textbf{55.1} & \textbf{68.1} \\ \hline
\end{tabular}
}
\caption {Performance comparison on document-level closed-domain event extraction (\%). Bold indicates the best performance.}
\label{tab:table3}
\end{table}

\section{Generality of MODEE in Closed-Domain Event Extraction}
To assess the generality of MODEE, we trained and evaluated our best-performing model, MODEE-Base, on DocEE \cite{tongDocEELargeScaleFinegrained2022}. We used DocEE because it is one of the largest manually annotated closed-domain event extraction datasets and follows the one-event-per-document paradigm. We trained MODEE-Base using inputs that consist of a document followed by the argument types associated with the main event it describes so that the model learns to generate the corresponding event arguments. We compared the performance of MODEE-Base against the state-of-the-art (SOTA) algorithms: BERT\_Seq \cite{duDocumentLevelEventRole2020}, MG-Reader \cite{duDocumentLevelEventRole2020}, Doc2EDAG \cite{zhengDoc2EDAGEndtoEndDocumentlevel2019}, BERT\_QA \cite{duEventExtractionAnswering2020}, and Ontology\_QA \cite{tongDocEELargeScaleFinegrained2022}. For a fair comparison, we adopted the same train/test split (22k train, 2.7k test) under normal-setting used in \citet{tongDocEELargeScaleFinegrained2022}. We trained MODEE-Base for 20 epochs using the configuration described in Section 4.2, except the input and output sequence lengths were limited to 1,024 tokens.

\setlength{\parskip}{0pt}Table \ref{tab:table3} presents the experimental results using precision ($P$), recall ($R$), and $F1$ scores computed with exact match (EM) and head noun phrase match (HM), following prior work \cite{tongDocEELargeScaleFinegrained2022}. For the SOTA algorithms, we use the reported performance from \citet{tongDocEELargeScaleFinegrained2022}. MODEE-Base has better performance than all SOTA algorithms for both EM and HM, despite being primarily designed for open-domain event extraction. One reason for lower accuracy of the existing SOTA algorithms is their limited ability to understand the document-level semantics of event related tokens within overall document context. In contrast, MODEE-Base, through its integration of graph- and text-based representations, more effectively utilizes the document-level context and semantic relationships of the event related tokens in a document. As a result, it achieves improved performance, and this experimental result demonstrates the generality and adaptability of MODEE.

\section{Conclusion}
In this research, we introduced MODEE, a novel GNN- and LLM-based open-domain event extraction approach that combines multiple modalities to model document-level reasoning, which is essential for event extraction but remains challenging for LLMs due to lost-in-the-middle phenomenon and attention dilution. Experimental results showed that MODEE outperformed strong baselines in both open-domain and closed-domain settings. Future work will extend MODEE to multi-document settings and incorporate additional modalities such as images, audio, and video.

\section*{Limitations}

We evaluate MODEE’s event extraction capability using news reports. While this domain is well suited for event 5Ws extraction, additional experiments on other domains such as social media, scientific articles, or financial documents are necessary to fully assess MODEE across diverse text genres. MODEE assumes a one-event-per-document setting. When a document reports multiple salient events, the model is not explicitly designed to extract all events and may focus on a single dominant event.

The document-level token graph is a critical component of MODEE as it enables the model to capture the global structure and semantic relationships among event-related tokens. However, constructing and processing a document-level token graph for a very long document introduces additional computational overhead, which may limit MODEE’s scalability without further optimization. Techniques such as graph sparsification or edge pruning could help reduce this complexity. Finally, MODEE uses T5 as its backbone due to its encoder-decoder architecture. Most modern LLMs are decoder-only models, and we have not evaluated MODEE with larger and more recent LLM backbones, which could potentially allow better utilization of the capabilities of modern LLMs.

\section*{Ethical Considerations}

Creating a large, manually annotated dataset requires substantial time and effort. We employed student annotators and compensated them at the basic minimum pay rate set by the university for student workers. Participation in the annotation process was voluntary, and to protect privacy and confidentiality, no personal or identifying information about the annotators is disclosed.

\bibliography{custom}

\appendix 
\section{Dataset Details} \label{appen1}
\begin{figure*}[t]
  \includegraphics[width=\linewidth]{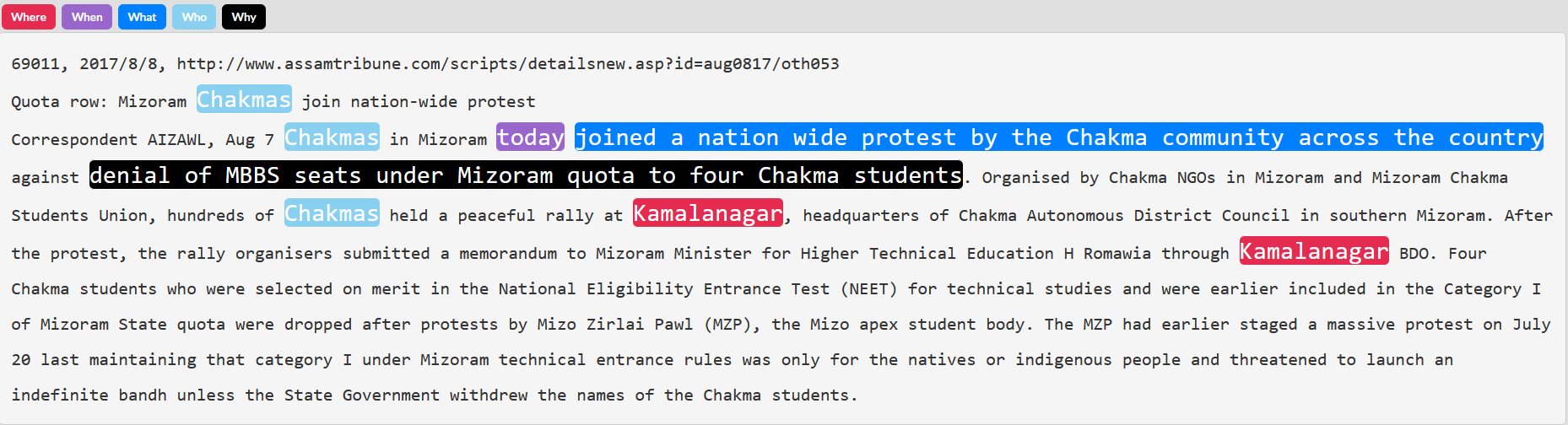}
  \caption{Example of a document from our dataset with the 5Ws for the main event annotated (highlighted).}
  \label{fig:figure2}
\end{figure*}

Figure \ref{fig:figure2}   shows an example of the main event 5Ws annotated in a document from our dataset. The dataset has an average document length of 127 words, with each document containing 4.1 of the five Ws on average. What is present in all documents, while Where, When, Who, and Why appear in 9,390 (93.9\%), 9,493 (94.9\%), 7,391 (73.9\%), and 4,451 (44.5\%) documents, respectively. This distribution highlights the relative sparsity of causal information (Why) in news reports, which makes identifying and extracting it complex. In terms of lexical diversity, the dataset contains 5,469 (58.2\%) unique Where values, 1,279 (13.5\%) When, 7,462 (74.6\%) What, 4,424 (59.9\%) Who, and 3,944 (88.6\%) Why. This indicates substantial variability, particularly for What and Why, making generalization more challenging. The average span lengths (in words) are 1.6 (Where), 1.0 (When), 5.8 (What), 2.3 (Who), and 7.4 (Why) and suggest that What and Why are typically longer, more descriptive, and semantically complex. Overall, the dataset exhibits high lexical diversity, substantial variation across values for the 5Ws, multi-word spans, and sparse causal information, which collectively make the task of event 5Ws extraction challenging.

\section{Prompt Examples} \label{appen2}
\begin{figure*}[t]
  \includegraphics[width=\linewidth]{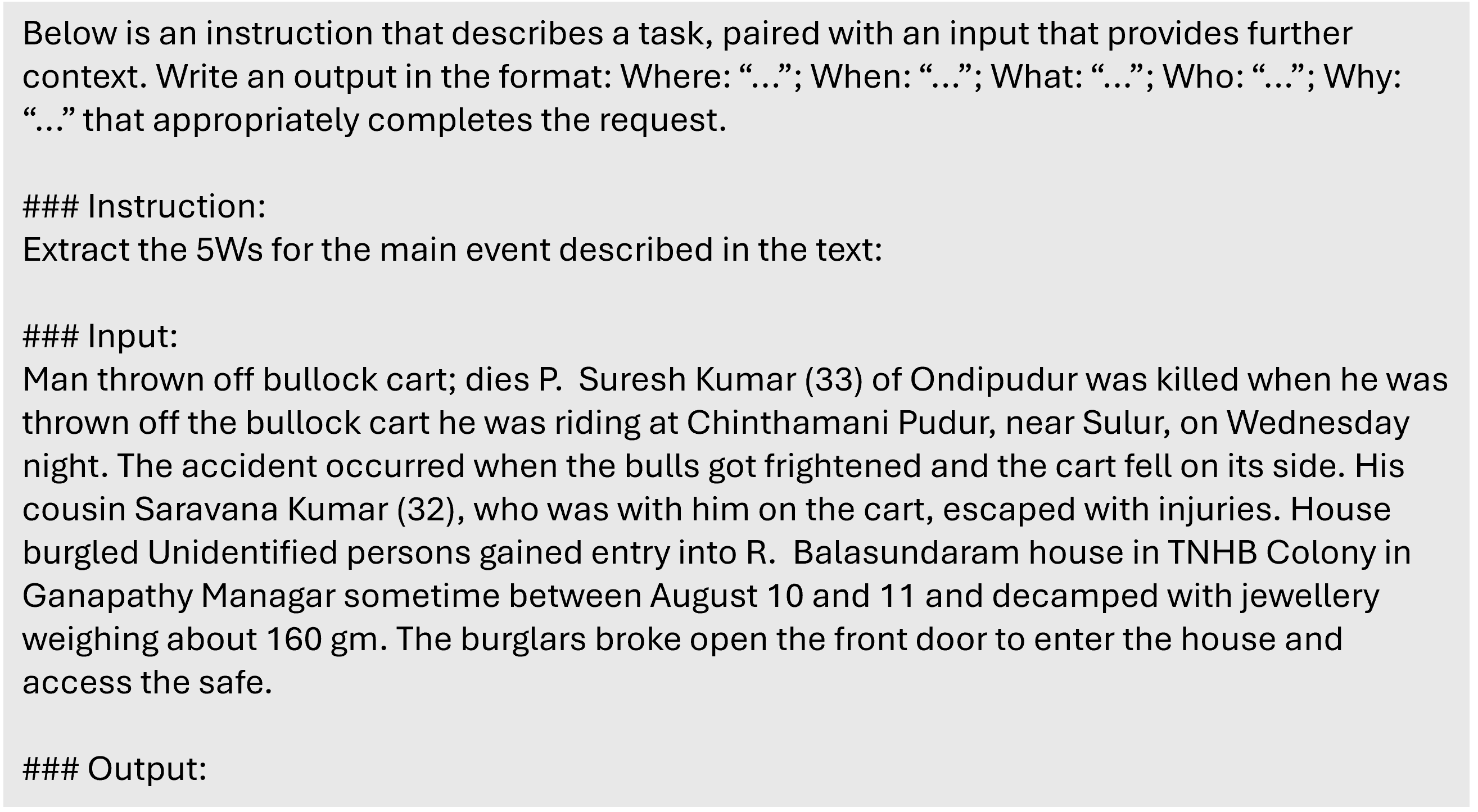}
  \caption{One-shot prompt example.}
  \label{fig:figure3}
\end{figure*}

\begin{figure*}[t]
  \includegraphics[width=\linewidth]{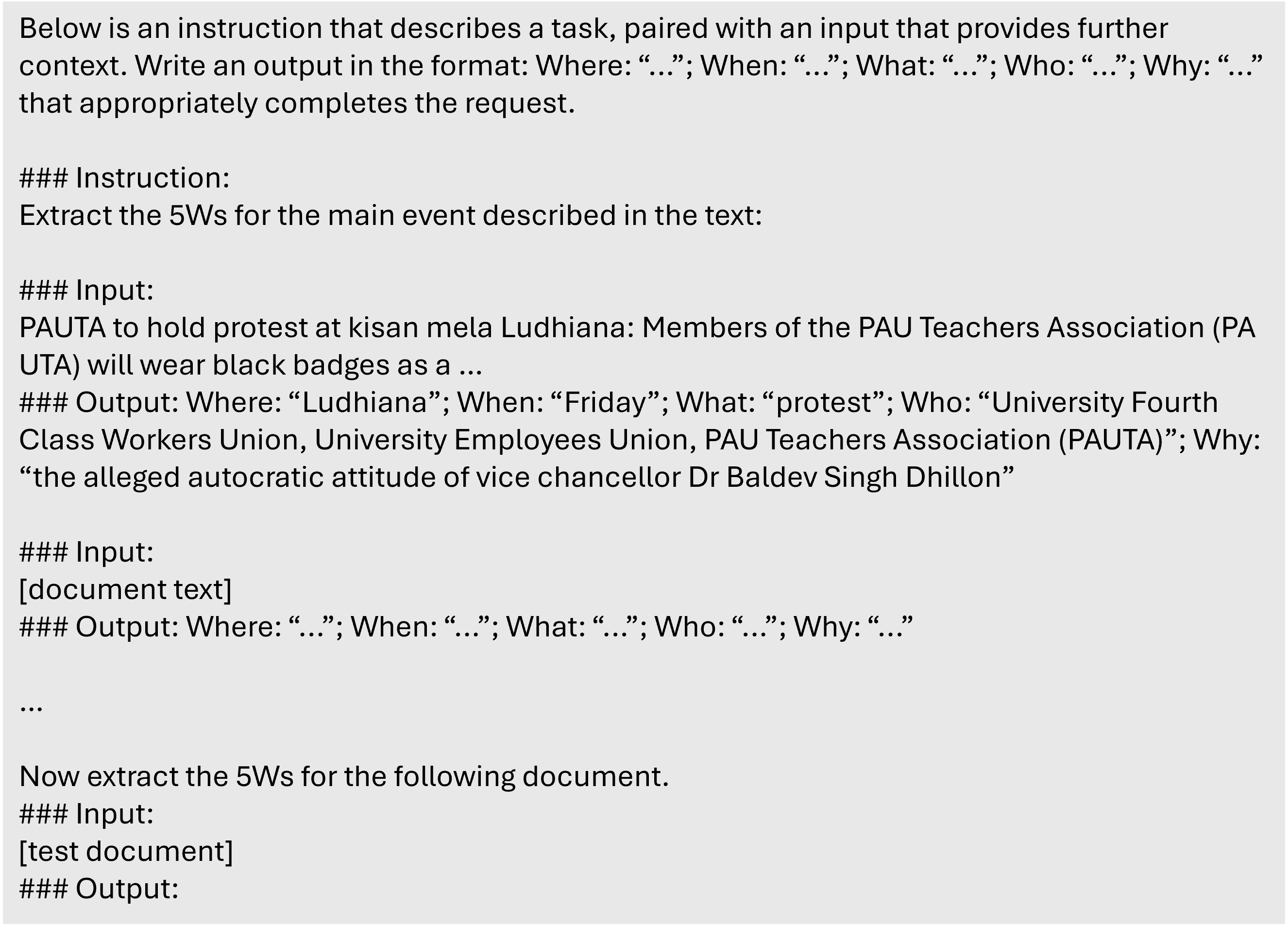}
  \caption{One-shot prompt example.}
  \label{fig:figure4}
\end{figure*}

We adopt Alpaca-style prompting for both zero- and five-shot settings, and use the same format for fine-tuning the LLMs in this study (see Figures \ref{fig:figure3} and \ref{fig:figure4}). Our focus was on comparing the performance of different models under a consistent setup. Although more sophisticated prompt engineering may further improve performance, such optimization is outside the scope of this work.

\section{Main Results (Full Tables)} \label{appen3}
Tables \ref{tab:my-table4}, \ref{tab:my-table5}, and \ref{tab:my-table6} report precision (P), recall (R), and F1 scores for each 5W class using Exact Match, ROUGE-L, and BERTScore. Models perform better on Where, When, and Who. Performance on What and especially Why is lower, which reflects their greater difficulty. Why is particularly challenging due to higher lexical variability and sparsity. Notably, MODEE achieves the strongest performance on Why, outperforming all baselines, including fine-tuned LLMs, despite its smaller size. This improvement likely stems from its integration of textual and graph-based representations for modeling document-level reasoning.

\begin{table*}[t]
\resizebox{\textwidth}{!}{
\begin{tabular}{l|ccc|ccc|ccc|ccc|ccc|ccc}
\hline
\multicolumn{1}{c|}{\multirow{2}{*}{\textbf{Models}}} & \multicolumn{3}{c|}{\textbf{Where}} & \multicolumn{3}{c|}{\textbf{When}} & \multicolumn{3}{c|}{\textbf{What}} & \multicolumn{3}{c|}{\textbf{Who}} & \multicolumn{3}{c|}{\textbf{Why}} & \multicolumn{3}{c}{\textbf{Overall}} \\ \cline{2-19} 
\multicolumn{1}{c|}{} & \multicolumn{1}{c|}{\textit{P}} & \multicolumn{1}{c|}{\textit{R}} & \textit{F1} & \multicolumn{1}{c|}{\textit{P}} & \multicolumn{1}{c|}{\textit{R}} & \textit{F1} & \multicolumn{1}{c|}{\textit{P}} & \multicolumn{1}{c|}{\textit{R}} & \textit{F1} & \multicolumn{1}{c|}{\textit{P}} & \multicolumn{1}{c|}{\textit{R}} & \textit{F1} & \multicolumn{1}{c|}{\textit{P}} & \multicolumn{1}{c|}{\textit{R}} & \textit{F1} & \multicolumn{1}{c|}{\textit{P}} & \multicolumn{1}{c}{\textit{R}} & \textit{F1} \\ \hline
\textbf{T5-Small (Fine-tuned)} & \multicolumn{1}{c|}{57.1} & \multicolumn{1}{c|}{57.9} & 57.5 & \multicolumn{1}{c|}{85.3} & \multicolumn{1}{c|}{76.7} & 80.7 & \multicolumn{1}{c|}{29.2} & \multicolumn{1}{c|}{29.2} & 29.2 & \multicolumn{1}{c|}{45.6} & \multicolumn{1}{c|}{40.0} & 42.6 & \multicolumn{1}{c|}{30.5} & \multicolumn{1}{c|}{17.4} & 22.2 & \multicolumn{1}{c|}{52.2} & \multicolumn{1}{c|}{47.5} & 49.8 \\ 
\textbf{T5-Base (Fine-tuned)} & \multicolumn{1}{c|}{60.8} & \multicolumn{1}{c|}{62.1} & 61.4 & \multicolumn{1}{c|}{85.7} & \multicolumn{1}{c|}{78.7} & 82.0 & \multicolumn{1}{c|}{30.8} & \multicolumn{1}{c|}{30.8} & 30.8 & \multicolumn{1}{c|}{54.0} & \multicolumn{1}{c|}{49.5} & 51.7 & \multicolumn{1}{c|}{31.7} & \multicolumn{1}{c|}{24.0} & 27.3 & \multicolumn{1}{c|}{54.9} & \multicolumn{1}{c|}{51.8} & 53.3 \\ 
\textbf{T5-Large (Fine-tuned)} & \multicolumn{1}{c|}{65.3} & \multicolumn{1}{c|}{66.8} & 66.0 & \multicolumn{1}{c|}{87.6} & \multicolumn{1}{c|}{79.2} & \textbf{83.2} & \multicolumn{1}{c|}{33.7} & \multicolumn{1}{c|}{33.6} & 33.6 & \multicolumn{1}{c|}{56.2} & \multicolumn{1}{c|}{53.7} & 54.9 & \multicolumn{1}{c|}{30.1} & \multicolumn{1}{c|}{24.0} & 26.7 & \multicolumn{1}{c|}{57.2} & \multicolumn{1}{c|}{54.4} & 55.8 \\ 
\textbf{Giveme5W1H} & \multicolumn{1}{c|}{21.9} & \multicolumn{1}{c|}{16.5} & 18.8 & \multicolumn{1}{c|}{60.7} & \multicolumn{1}{c|}{47.1} & 53.0 & \multicolumn{1}{c|}{1.5} & \multicolumn{1}{c|}{1.5} & 1.5 & \multicolumn{1}{c|}{6.4} & \multicolumn{1}{c|}{8.3} & 7.2 & \multicolumn{1}{c|}{0.1} & \multicolumn{1}{c|}{0.2} & 0.1 & \multicolumn{1}{c|}{15.6} & \multicolumn{1}{c|}{16.6} & 16.1 \\ 
\textbf{Llama 8B (0-shot)} & \multicolumn{1}{c|}{11.0} & \multicolumn{1}{c|}{11.6} & 11.3 & \multicolumn{1}{c|}{21.0} & \multicolumn{1}{c|}{19.3} & 20.1 & \multicolumn{1}{c|}{3.7} & \multicolumn{1}{c|}{3.2} & 3.4 & \multicolumn{1}{c|}{9.9} & \multicolumn{1}{c|}{11.2} & 10.5 & \multicolumn{1}{c|}{3.0} & \multicolumn{1}{c|}{5.5} & 3.9 & \multicolumn{1}{c|}{9.8} & \multicolumn{1}{c|}{10.6} & 10.2 \\ 
\textbf{Llama 8B (5-shot)} & \multicolumn{1}{c|}{11.1} & \multicolumn{1}{c|}{11.7} & 11.4 & \multicolumn{1}{c|}{20.0} & \multicolumn{1}{c|}{18.1} & 19.0 & \multicolumn{1}{c|}{4.2} & \multicolumn{1}{c|}{3.6} & 3.9 & \multicolumn{1}{c|}{10.2} & \multicolumn{1}{c|}{11.5} & 10.8 & \multicolumn{1}{c|}{2.1} & \multicolumn{1}{c|}{3.7} & 2.7 & \multicolumn{1}{c|}{9.6} & \multicolumn{1}{c|}{10.3} & 10.0 \\ 
\textbf{Llama 70B (0-shot)} & \multicolumn{1}{c|}{15.4} & \multicolumn{1}{c|}{15.8} & 15.6 & \multicolumn{1}{c|}{30.2} & \multicolumn{1}{c|}{24.4} & 27.0 & \multicolumn{1}{c|}{6.0} & \multicolumn{1}{c|}{4.4} & 5.1 & \multicolumn{1}{c|}{14.7} & \multicolumn{1}{c|}{14.2} & 14.4 & \multicolumn{1}{c|}{2.8} & \multicolumn{1}{c|}{4.4} & 3.4 & \multicolumn{1}{c|}{14.1} & \multicolumn{1}{c|}{13.4} & 13.8 \\ 
\textbf{Llama 70B (5-shot)} & \multicolumn{1}{c|}{16.0} & \multicolumn{1}{c|}{16.3} & 16.1 & \multicolumn{1}{c|}{31.1} & \multicolumn{1}{c|}{24.8} & 27.6 & \multicolumn{1}{c|}{6.7} & \multicolumn{1}{c|}{4.9} & 5.7 & \multicolumn{1}{c|}{13.9} & \multicolumn{1}{c|}{13.4} & 13.6 & \multicolumn{1}{c|}{3.0} & \multicolumn{1}{c|}{4.6} & 3.6 & \multicolumn{1}{c|}{14.4} & \multicolumn{1}{c|}{13.7} & 14.0 \\ 
\textbf{Qwen 8B (0-shot)} & \multicolumn{1}{c|}{26.0} & \multicolumn{1}{c|}{27.4} & 26.7 & \multicolumn{1}{c|}{38.0} & \multicolumn{1}{c|}{39.7} & 38.8 & \multicolumn{1}{c|}{5.5} & \multicolumn{1}{c|}{5.5} & 5.5 & \multicolumn{1}{c|}{8.5} & \multicolumn{1}{c|}{11.4} & 9.7 & \multicolumn{1}{c|}{0.9} & \multicolumn{1}{c|}{2.0} & 1.2 & \multicolumn{1}{c|}{15.8} & \multicolumn{1}{c|}{19.2} & 17.3 \\ 
\textbf{Qwen 8B (5-shot)} & \multicolumn{1}{c|}{26.8} & \multicolumn{1}{c|}{28.3} & 27.5 & \multicolumn{1}{c|}{37.6} & \multicolumn{1}{c|}{39.5} & 38.6 & \multicolumn{1}{c|}{5.3} & \multicolumn{1}{c|}{5.3} & 5.3 & \multicolumn{1}{c|}{8.3} & \multicolumn{1}{c|}{11.1} & 9.5 & \multicolumn{1}{c|}{1.0} & \multicolumn{1}{c|}{2.2} & 1.4 & \multicolumn{1}{c|}{15.8} & \multicolumn{1}{c|}{19.3} & 17.4 \\ 
\textbf{Qwen 32B (0-shot)} & \multicolumn{1}{c|}{15.3} & \multicolumn{1}{c|}{15.5} & 15.4 & \multicolumn{1}{c|}{30.7} & \multicolumn{1}{c|}{30.9} & 30.8 & \multicolumn{1}{c|}{0.8} & \multicolumn{1}{c|}{0.8} & 0.8 & \multicolumn{1}{c|}{2.4} & \multicolumn{1}{c|}{3.1} & 2.7 & \multicolumn{1}{c|}{0.4} & \multicolumn{1}{c|}{0.9} & 0.6 & \multicolumn{1}{c|}{9.9} & \multicolumn{1}{c|}{11.6} & 10.7 \\ 
\textbf{Qwen 32B (5-shot)} & \multicolumn{1}{c|}{13.2} & \multicolumn{1}{c|}{13.3} & 13.2 & \multicolumn{1}{c|}{26.9} & \multicolumn{1}{c|}{27.3} & 27.1 & \multicolumn{1}{c|}{1.6} & \multicolumn{1}{c|}{1.5} & 1.5 & \multicolumn{1}{c|}{2.4} & \multicolumn{1}{c|}{3.1} & 2.7 & \multicolumn{1}{c|}{0.6} & \multicolumn{1}{c|}{1.3} & 0.9 & \multicolumn{1}{c|}{9.0} & \multicolumn{1}{c|}{10.5} & 9.7 \\ 
\textbf{Mistral 7B (0-shot)} & \multicolumn{1}{c|}{18.8} & \multicolumn{1}{c|}{19.8} & 19.3 & \multicolumn{1}{c|}{34.0} & \multicolumn{1}{c|}{35.5} & 34.7 & \multicolumn{1}{c|}{6.9} & \multicolumn{1}{c|}{6.9} & 6.9 & \multicolumn{1}{c|}{15.0} & \multicolumn{1}{c|}{19.9} & 17.1 & \multicolumn{1}{c|}{2.8} & \multicolumn{1}{c|}{6.2} & 3.9 & \multicolumn{1}{c|}{15.5} & \multicolumn{1}{c|}{18.8} & 17.0 \\ 
\textbf{Mistral 7B (5-shot)} & \multicolumn{1}{c|}{18.1} & \multicolumn{1}{c|}{19.0} & 18.6 & \multicolumn{1}{c|}{34.0} & \multicolumn{1}{c|}{35.4} & 34.7 & \multicolumn{1}{c|}{7.0} & \multicolumn{1}{c|}{7.0} & 7.0 & \multicolumn{1}{c|}{14.9} & \multicolumn{1}{c|}{19.8} & 17.0 & \multicolumn{1}{c|}{2.8} & \multicolumn{1}{c|}{6.2} & 3.9 & \multicolumn{1}{c|}{15.4} & \multicolumn{1}{c|}{18.6} & 16.8 \\ 
\textbf{Llama 1B (Fine-tuned)} & \multicolumn{1}{c|}{64.4} & \multicolumn{1}{c|}{64.2} & 64.3 & \multicolumn{1}{c|}{78.8} & \multicolumn{1}{c|}{78.2} & 78.5 & \multicolumn{1}{c|}{38.4} & \multicolumn{1}{c|}{38.4} & 38.4 & \multicolumn{1}{c|}{61.5} & \multicolumn{1}{c|}{56.2} & 58.7 & \multicolumn{1}{c|}{34.9} & \multicolumn{1}{c|}{24.9} & 29.0 & \multicolumn{1}{c|}{58.2} & \multicolumn{1}{c|}{55.3} & 56.8 \\ 
\textbf{Qwen 0.6B (Fine-tuned)} & \multicolumn{1}{c|}{64.5} & \multicolumn{1}{c|}{64.1} & 64.3 & \multicolumn{1}{c|}{77.6} & \multicolumn{1}{c|}{77.8} & 77.7 & \multicolumn{1}{c|}{39.2} & \multicolumn{1}{c|}{39.2} & \textbf{39.2} & \multicolumn{1}{c|}{60.6} & \multicolumn{1}{c|}{64.5} & \textbf{62.5} & \multicolumn{1}{c|}{27.2} & \multicolumn{1}{c|}{31.9} & 29.4 & \multicolumn{1}{c|}{56.0} & \multicolumn{1}{c|}{57.7} & 56.9 \\ 
\textbf{MODEE-Small} & \multicolumn{1}{c|}{57.7} & \multicolumn{1}{c|}{61.0} & 59.3 & \multicolumn{1}{c|}{86.0} & \multicolumn{1}{c|}{77.6} & 81.6 & \multicolumn{1}{c|}{35.0} & \multicolumn{1}{c|}{35.0} & 35.0 & \multicolumn{1}{c|}{52.4} & \multicolumn{1}{c|}{47.8} & 50.0 & \multicolumn{1}{c|}{29.0} & \multicolumn{1}{c|}{21.4} & 24.6 & \multicolumn{1}{c|}{54.7} & \multicolumn{1}{c|}{51.7} & 53.2 \\ 
\textbf{MODEE-Base} & \multicolumn{1}{c|}{67.0} & \multicolumn{1}{c|}{68.0} & \textbf{67.5} & \multicolumn{1}{c|}{85.7} & \multicolumn{1}{c|}{79.4} & 82.4 & \multicolumn{1}{c|}{35.7} & \multicolumn{1}{c|}{35.7} & 35.7 & \multicolumn{1}{c|}{59.1} & \multicolumn{1}{c|}{56.9} & 58.0 & \multicolumn{1}{c|}{35.7} & \multicolumn{1}{c|}{31.5} & \textbf{33.5} & \multicolumn{1}{c|}{58.7} & \multicolumn{1}{c|}{56.7} & \textbf{57.7} \\ \hline
\end{tabular}
}
\caption{Performance comparison on event extraction (\%) using Exact Match. Bold indicates the best performance.}
\label{tab:my-table4}
\end{table*}

\begin{table*}[t]
\resizebox{\textwidth}{!}{
\begin{tabular}{l|lll|lll|lll|lll|lll|lll}
\hline
\multicolumn{1}{c|}{\multirow{2}{*}{\textbf{Models}}} & \multicolumn{3}{c|}{\textbf{Where}} & \multicolumn{3}{c|}{\textbf{When}} & \multicolumn{3}{c|}{\textbf{What}} & \multicolumn{3}{c|}{\textbf{Who}} & \multicolumn{3}{c|}{\textbf{Why}} & \multicolumn{3}{c}{\textbf{Overall}} \\ \cline{2-19} 
\multicolumn{1}{c|}{} & \multicolumn{1}{c|}{\textit{P}} & \multicolumn{1}{c|}{\textit{R}} & \multicolumn{1}{c|}{\textit{F1}} & \multicolumn{1}{c|}{\textit{P}} & \multicolumn{1}{c|}{\textit{R}} & \multicolumn{1}{c|}{\textit{F1}} & \multicolumn{1}{c|}{\textit{P}} & \multicolumn{1}{c|}{\textit{R}} & \multicolumn{1}{c|}{\textit{F1}} & \multicolumn{1}{c|}{\textit{P}} & \multicolumn{1}{c|}{\textit{R}} & \multicolumn{1}{c|}{\textit{F1}} & \multicolumn{1}{c|}{\textit{P}} & \multicolumn{1}{c|}{\textit{R}} & \multicolumn{1}{c|}{\textit{F1}} & \multicolumn{1}{c|}{\textit{P}} & \multicolumn{1}{c|}{\textit{R}} & \multicolumn{1}{c}{\textit{F1}} \\ \hline
\textbf{T5-Small (Fine-tuned)} & \multicolumn{1}{l|}{65.5} & \multicolumn{1}{l|}{66.9} & 65.3 & \multicolumn{1}{l|}{87.1} & \multicolumn{1}{l|}{87.0} & 87.0 & \multicolumn{1}{l|}{60.8} & \multicolumn{1}{l|}{60.4} & 56.5 & \multicolumn{1}{l|}{68.0} & \multicolumn{1}{l|}{64.7} & 64.5 & \multicolumn{1}{l|}{67.9} & \multicolumn{1}{l|}{71.4} & 66.1 & \multicolumn{1}{l|}{69.8} & \multicolumn{1}{l|}{69.7} & 67.8 \\ 
\textbf{T5-Base (Fine-tuned)} & \multicolumn{1}{l|}{67.8} & \multicolumn{1}{l|}{69.2} & 67.8 & \multicolumn{1}{l|}{87.9} & \multicolumn{1}{l|}{87.8} & 87.8 & \multicolumn{1}{l|}{61.4} & \multicolumn{1}{l|}{63.6} & 58.5 & \multicolumn{1}{l|}{73.0} & \multicolumn{1}{l|}{72.2} & 71.0 & \multicolumn{1}{l|}{69.3} & \multicolumn{1}{l|}{73.3} & 67.7 & \multicolumn{1}{l|}{71.7} & \multicolumn{1}{l|}{72.8} & 70.5 \\ 
\textbf{T5-Large (Fine-tuned)} & \multicolumn{1}{l|}{73.5} & \multicolumn{1}{l|}{75.0} & 73.4 & \multicolumn{1}{l|}{89.5} & \multicolumn{1}{l|}{89.4} & \textbf{89.4} & \multicolumn{1}{l|}{65.9} & \multicolumn{1}{l|}{64.5} & 60.7 & \multicolumn{1}{l|}{76.2} & \multicolumn{1}{l|}{76.1} & 74.4 & \multicolumn{1}{l|}{67.7} & \multicolumn{1}{l|}{67.2} & 63.3 & \multicolumn{1}{l|}{75.2} & \multicolumn{1}{l|}{75.1} & 73.0 \\ 
\textbf{Giveme5W1H} & \multicolumn{1}{l|}{31.9} & \multicolumn{1}{l|}{28.6} & 29.1 & \multicolumn{1}{l|}{64.4} & \multicolumn{1}{l|}{66.6} & 64.9 & \multicolumn{1}{l|}{21.4} & \multicolumn{1}{l|}{21.9} & 18.2 & \multicolumn{1}{l|}{22.4} & \multicolumn{1}{l|}{28.1} & 21.5 & \multicolumn{1}{l|}{11.7} & \multicolumn{1}{l|}{5.6} & 5.3 & \multicolumn{1}{l|}{31.1} & \multicolumn{1}{l|}{31.4} & 28.8 \\ 
\textbf{Llama 8B (0-shot)} & \multicolumn{1}{l|}{25.2} & \multicolumn{1}{l|}{58.7} & 30.0 & \multicolumn{1}{l|}{32.6} & \multicolumn{1}{l|}{54.4} & 36.9 & \multicolumn{1}{l|}{25.6} & \multicolumn{1}{l|}{45.3} & 25.3 & \multicolumn{1}{l|}{27.8} & \multicolumn{1}{l|}{53.8} & 31.6 & \multicolumn{1}{l|}{27.3} & \multicolumn{1}{l|}{48.9} & 30.5 & \multicolumn{1}{l|}{27.7} & \multicolumn{1}{l|}{52.7} & 30.8 \\ 
\textbf{Llama 8B (5-shot)} & \multicolumn{1}{l|}{24.7} & \multicolumn{1}{l|}{57.1} & 29.2 & \multicolumn{1}{l|}{30.6} & \multicolumn{1}{l|}{52.2} & 34.8 & \multicolumn{1}{l|}{27.4} & \multicolumn{1}{l|}{46.6} & 26.6 & \multicolumn{1}{l|}{27.6} & \multicolumn{1}{l|}{54.0} & 31.4 & \multicolumn{1}{l|}{24.4} & \multicolumn{1}{l|}{45.7} & 27.5 & \multicolumn{1}{l|}{27.1} & \multicolumn{1}{l|}{51.8} & 30.1 \\ 
\textbf{Llama 70B (0-shot)} & \multicolumn{1}{l|}{24.0} & \multicolumn{1}{l|}{49.6} & 26.3 & \multicolumn{1}{l|}{40.2} & \multicolumn{1}{l|}{55.1} & 42.9 & \multicolumn{1}{l|}{33.3} & \multicolumn{1}{l|}{35.8} & 27.6 & \multicolumn{1}{l|}{31.7} & \multicolumn{1}{l|}{42.9} & 33.1 & \multicolumn{1}{l|}{25.4} & \multicolumn{1}{l|}{34.9} & 25.1 & \multicolumn{1}{l|}{31.2} & \multicolumn{1}{l|}{45.2} & 31.4 \\ 
\textbf{Llama 70B (5-shot)} & \multicolumn{1}{l|}{25.3} & \multicolumn{1}{l|}{53.0} & 28.1 & \multicolumn{1}{l|}{41.6} & \multicolumn{1}{l|}{56.4} & 44.4 & \multicolumn{1}{l|}{31.4} & \multicolumn{1}{l|}{32.6} & 25.4 & \multicolumn{1}{l|}{29.9} & \multicolumn{1}{l|}{39.4} & 31.1 & \multicolumn{1}{l|}{26.7} & \multicolumn{1}{l|}{35.7} & 26.4 & \multicolumn{1}{l|}{31.3} & \multicolumn{1}{l|}{45.2} & 31.5 \\ 
\textbf{Qwen 8B (0-shot)} & \multicolumn{1}{l|}{38.7} & \multicolumn{1}{l|}{51.4} & 40.8 & \multicolumn{1}{l|}{49.0} & \multicolumn{1}{l|}{64.0} & 52.6 & \multicolumn{1}{l|}{34.6} & \multicolumn{1}{l|}{53.2} & 36.6 & \multicolumn{1}{l|}{29.2} & \multicolumn{1}{l|}{61.3} & 35.5 & \multicolumn{1}{l|}{27.2} & \multicolumn{1}{l|}{45.5} & 31.1 & \multicolumn{1}{l|}{37.1} & \multicolumn{1}{l|}{55.9} & 40.5 \\ 
\textbf{Qwen 8B (5-shot)} & \multicolumn{1}{l|}{39.2} & \multicolumn{1}{l|}{52.9} & 41.6 & \multicolumn{1}{l|}{48.9} & \multicolumn{1}{l|}{64.4} & 52.4 & \multicolumn{1}{l|}{35.2} & \multicolumn{1}{l|}{53.5} & 36.6 & \multicolumn{1}{l|}{28.8} & \multicolumn{1}{l|}{61.1} & 35.2 & \multicolumn{1}{l|}{26.7} & \multicolumn{1}{l|}{45.7} & 30.4 & \multicolumn{1}{l|}{37.2} & \multicolumn{1}{l|}{56.4} & 40.5 \\ 
\textbf{Qwen 32B (0-shot)} & \multicolumn{1}{l|}{37.0} & \multicolumn{1}{l|}{77.8} & 45.5 & \multicolumn{1}{l|}{43.4} & \multicolumn{1}{l|}{68.9} & 48.5 & \multicolumn{1}{l|}{24.0} & \multicolumn{1}{l|}{62.5} & 30.8 & \multicolumn{1}{l|}{20.9} & \multicolumn{1}{l|}{72.4} & 29.0 & \multicolumn{1}{l|}{18.4} & \multicolumn{1}{l|}{50.1} & 24.0 & \multicolumn{1}{l|}{30.3} & \multicolumn{1}{l|}{68.0} & 37.2 \\ 
\textbf{Qwen 32B (5-shot)} & \multicolumn{1}{l|}{35.4} & \multicolumn{1}{l|}{77.5} & 44.3 & \multicolumn{1}{l|}{40.3} & \multicolumn{1}{l|}{68.6} & 45.9 & \multicolumn{1}{l|}{24.5} & \multicolumn{1}{l|}{62.0} & 30.7 & \multicolumn{1}{l|}{21.1} & \multicolumn{1}{l|}{71.3} & 29.4 & \multicolumn{1}{l|}{18.4} & \multicolumn{1}{l|}{51.6} & 24.3 & \multicolumn{1}{l|}{29.4} & \multicolumn{1}{l|}{67.7} & 36.5 \\ 
\textbf{Mistral 7B (0-shot)} & \multicolumn{1}{l|}{39.0} & \multicolumn{1}{l|}{62.1} & 44.2 & \multicolumn{1}{l|}{46.2} & \multicolumn{1}{l|}{59.5} & 50.0 & \multicolumn{1}{l|}{33.6} & \multicolumn{1}{l|}{40.3} & 29.1 & \multicolumn{1}{l|}{31.7} & \multicolumn{1}{l|}{46.6} & 34.3 & \multicolumn{1}{l|}{28.6} & \multicolumn{1}{l|}{44.7} & 31.1 & \multicolumn{1}{l|}{36.9} & \multicolumn{1}{l|}{51.4} & 38.6 \\ 
\textbf{Mistral 7B (5-shot)} & \multicolumn{1}{l|}{38.3} & \multicolumn{1}{l|}{61.6} & 43.6 & \multicolumn{1}{l|}{46.6} & \multicolumn{1}{l|}{60.2} & 50.5 & \multicolumn{1}{l|}{33.3} & \multicolumn{1}{l|}{40.2} & 29.1 & \multicolumn{1}{l|}{31.4} & \multicolumn{1}{l|}{46.0} & 34.0 & \multicolumn{1}{l|}{28.2} & \multicolumn{1}{l|}{43.3} & 30.6 & \multicolumn{1}{l|}{36.6} & \multicolumn{1}{l|}{51.2} & 38.5 \\ 
\textbf{Llama 1B (Fine-tuned)} & \multicolumn{1}{l|}{71.2} & \multicolumn{1}{l|}{71.0} & 70.5 & \multicolumn{1}{l|}{81.9} & \multicolumn{1}{l|}{81.8} & 81.8 & \multicolumn{1}{l|}{65.1} & \multicolumn{1}{l|}{63.3} & 60.3 & \multicolumn{1}{l|}{78.8} & \multicolumn{1}{l|}{77.2} & 76.6 & \multicolumn{1}{l|}{72.9} & \multicolumn{1}{l|}{69.7} & 67.0 & \multicolumn{1}{l|}{73.6} & \multicolumn{1}{l|}{72.5} & 71.3 \\ 
\textbf{Qwen 0.6B (Fine-tuned)} & \multicolumn{1}{l|}{71.5} & \multicolumn{1}{l|}{71.0} & 70.6 & \multicolumn{1}{l|}{80.7} & \multicolumn{1}{l|}{80.6} & 80.6 & \multicolumn{1}{l|}{64.2} & \multicolumn{1}{l|}{66.5} & 60.9 & \multicolumn{1}{l|}{79.8} & \multicolumn{1}{l|}{78.8} & \textbf{77.9} & \multicolumn{1}{l|}{66.6} & \multicolumn{1}{l|}{67.2} & 62.8 & \multicolumn{1}{l|}{72.8} & \multicolumn{1}{l|}{73.2} & 71.1 \\ 
\textbf{MODEE-Small} & \multicolumn{1}{l|}{65.6} & \multicolumn{1}{l|}{66.3} & 65.2 & \multicolumn{1}{l|}{88.3} & \multicolumn{1}{l|}{88.2} & 88.2 & \multicolumn{1}{l|}{63.2} & \multicolumn{1}{l|}{66.7} & 60.8 & \multicolumn{1}{l|}{71.8} & \multicolumn{1}{l|}{71.4} & 70.1 & \multicolumn{1}{l|}{66.1} & \multicolumn{1}{l|}{74.0} & 65.8 & \multicolumn{1}{l|}{71.2} & \multicolumn{1}{l|}{72.8} & 70.1 \\ 
\textbf{MODEE-Base} & \multicolumn{1}{l|}{74.4} & \multicolumn{1}{l|}{75.1} & \textbf{74.0} & \multicolumn{1}{l|}{87.8} & \multicolumn{1}{l|}{87.7} & 87.7 & \multicolumn{1}{l|}{64.3} & \multicolumn{1}{l|}{65.9} & \textbf{61.1} & \multicolumn{1}{l|}{78.4} & \multicolumn{1}{l|}{78.0} & 76.6 & \multicolumn{1}{l|}{70.3} & \multicolumn{1}{l|}{73.3} & \textbf{68.2} & \multicolumn{1}{l|}{75.1} & \multicolumn{1}{l|}{75.9} & \textbf{73.7} \\ \hline
\end{tabular}
}
\caption{Performance comparison on event extraction (\%) using ROUGE-L. Bold indicates the best performance.}
\label{tab:my-table5}
\end{table*}

\begin{table*}[t]
\resizebox{\textwidth}{!}{
\begin{tabular}{l|lll|lll|lll|lll|lll|lll}
\hline
\multicolumn{1}{c|}{\multirow{2}{*}{\textbf{Models}}} & \multicolumn{3}{c|}{\textbf{Where}} & \multicolumn{3}{c|}{\textbf{When}} & \multicolumn{3}{c|}{\textbf{What}} & \multicolumn{3}{c|}{\textbf{Who}} & \multicolumn{3}{c|}{\textbf{Why}} & \multicolumn{3}{c}{\textbf{Overall}} \\ \cline{2-19} 
\multicolumn{1}{c|}{} & \multicolumn{1}{c|}{\textit{P}} & \multicolumn{1}{c|}{\textit{R}} & \multicolumn{1}{c|}{\textit{F1}} & \multicolumn{1}{c|}{\textit{P}} & \multicolumn{1}{c|}{\textit{R}} & \multicolumn{1}{c|}{\textit{F1}} & \multicolumn{1}{c|}{\textit{P}} & \multicolumn{1}{c|}{\textit{R}} & \multicolumn{1}{c|}{\textit{F1}} & \multicolumn{1}{c|}{\textit{P}} & \multicolumn{1}{c|}{\textit{R}} & \multicolumn{1}{c|}{\textit{F1}} & \multicolumn{1}{c|}{\textit{P}} & \multicolumn{1}{c|}{\textit{R}} & \multicolumn{1}{c|}{\textit{F1}} & \multicolumn{1}{c|}{\textit{P}} & \multicolumn{1}{c|}{\textit{R}} & \multicolumn{1}{c}{\textit{F1}} \\ \hline
\textbf{T5-Small (Fine-tuned)} & \multicolumn{1}{l|}{92.4} & \multicolumn{1}{l|}{92.6} & 92.5 & \multicolumn{1}{l|}{98.3} & \multicolumn{1}{l|}{98.2} & 98.2 & \multicolumn{1}{l|}{91.2} & \multicolumn{1}{l|}{91.4} & 91.2 & \multicolumn{1}{l|}{92.8} & \multicolumn{1}{l|}{92.3} & 92.4 & \multicolumn{1}{l|}{92.5} & \multicolumn{1}{l|}{93.5} & 93.0 & \multicolumn{1}{l|}{93.5} & \multicolumn{1}{l|}{93.6} & 93.5 \\ 
\textbf{T5-Base (Fine-tuned)} & \multicolumn{1}{l|}{93.0} & \multicolumn{1}{l|}{93.2} & 93.1 & \multicolumn{1}{l|}{98.3} & \multicolumn{1}{l|}{98.2} & 98.3 & \multicolumn{1}{l|}{91.7} & \multicolumn{1}{l|}{92.2} & 91.9 & \multicolumn{1}{l|}{93.8} & \multicolumn{1}{l|}{93.6} & 93.6 & \multicolumn{1}{l|}{93.3} & \multicolumn{1}{l|}{93.9} & \textbf{93.6} & \multicolumn{1}{l|}{94.1} & \multicolumn{1}{l|}{94.2} & 94.1 \\ 
\textbf{T5-Large (Fine-tuned)} & \multicolumn{1}{l|}{94.2} & \multicolumn{1}{l|}{94.4} & 94.2 & \multicolumn{1}{l|}{98.5} & \multicolumn{1}{l|}{98.4} & \textbf{98.4} & \multicolumn{1}{l|}{92.0} & \multicolumn{1}{l|}{92.0} & 92.0 & \multicolumn{1}{l|}{94.4} & \multicolumn{1}{l|}{94.4} & 94.4 & \multicolumn{1}{l|}{92.5} & \multicolumn{1}{l|}{92.5} & 92.5 & \multicolumn{1}{l|}{94.5} & \multicolumn{1}{l|}{94.5} & 94.5 \\ 
\textbf{Giveme5W1H} & \multicolumn{1}{l|}{84.6} & \multicolumn{1}{l|}{84.4} & 84.4 & \multicolumn{1}{l|}{94.3} & \multicolumn{1}{l|}{94.8} & 94.5 & \multicolumn{1}{l|}{84.7} & \multicolumn{1}{l|}{85.4} & 85.0 & \multicolumn{1}{l|}{85.5} & \multicolumn{1}{l|}{83.8} & 84.4 & \multicolumn{1}{l|}{83.3} & \multicolumn{1}{l|}{81.6} & 82.4 & \multicolumn{1}{l|}{86.6} & \multicolumn{1}{l|}{86.3} & 86.3 \\ 
\textbf{Llama 8B (0-shot)} & \multicolumn{1}{l|}{82.1} & \multicolumn{1}{l|}{86.8} & 84.3 & \multicolumn{1}{l|}{86.6} & \multicolumn{1}{l|}{90.1} & 88.3 & \multicolumn{1}{l|}{83.5} & \multicolumn{1}{l|}{87.1} & 85.2 & \multicolumn{1}{l|}{83.9} & \multicolumn{1}{l|}{87.8} & 85.7 & \multicolumn{1}{l|}{84.7} & \multicolumn{1}{l|}{87.8} & 86.1 & \multicolumn{1}{l|}{84.1} & \multicolumn{1}{l|}{87.9} & 85.8 \\ 
\textbf{Llama 8B (5-shot)} & \multicolumn{1}{l|}{82.2} & \multicolumn{1}{l|}{86.9} & 84.4 & \multicolumn{1}{l|}{86.4} & \multicolumn{1}{l|}{89.9} & 88.0 & \multicolumn{1}{l|}{83.7} & \multicolumn{1}{l|}{87.3} & 85.4 & \multicolumn{1}{l|}{83.6} & \multicolumn{1}{l|}{87.5} & 85.4 & \multicolumn{1}{l|}{84.1} & \multicolumn{1}{l|}{87.4} & 85.7 & \multicolumn{1}{l|}{83.9} & \multicolumn{1}{l|}{87.8} & 85.7 \\ 
\textbf{Llama 70B (0-shot)} & \multicolumn{1}{l|}{82.7} & \multicolumn{1}{l|}{85.9} & 84.1 & \multicolumn{1}{l|}{88.6} & \multicolumn{1}{l|}{90.9} & 89.7 & \multicolumn{1}{l|}{85.4} & \multicolumn{1}{l|}{86.6} & 85.9 & \multicolumn{1}{l|}{84.7} & \multicolumn{1}{l|}{87.5} & 86.0 & \multicolumn{1}{l|}{84.9} & \multicolumn{1}{l|}{86.5} & 85.6 & \multicolumn{1}{l|}{85.2} & \multicolumn{1}{l|}{87.5} & 86.2 \\ 
\textbf{Llama 70B (5-shot)} & \multicolumn{1}{l|}{83.3} & \multicolumn{1}{l|}{86.4} & 84.7 & \multicolumn{1}{l|}{89.0} & \multicolumn{1}{l|}{91.3} & 90.1 & \multicolumn{1}{l|}{85.0} & \multicolumn{1}{l|}{86.2} & 85.5 & \multicolumn{1}{l|}{84.3} & \multicolumn{1}{l|}{87.0} & 85.5 & \multicolumn{1}{l|}{85.2} & \multicolumn{1}{l|}{86.7} & 85.9 & \multicolumn{1}{l|}{85.3} & \multicolumn{1}{l|}{87.6} & 86.3 \\ 
\textbf{Qwen 8B (0-shot)} & \multicolumn{1}{l|}{85.9} & \multicolumn{1}{l|}{87.7} & 86.8 & \multicolumn{1}{l|}{90.7} & \multicolumn{1}{l|}{92.8} & 91.7 & \multicolumn{1}{l|}{86.4} & \multicolumn{1}{l|}{88.6} & 87.4 & \multicolumn{1}{l|}{83.4} & \multicolumn{1}{l|}{89.0} & 86.0 & \multicolumn{1}{l|}{86.1} & \multicolumn{1}{l|}{88.1} & 87.1 & \multicolumn{1}{l|}{86.7} & \multicolumn{1}{l|}{89.4} & 88.0 \\ 
\textbf{Qwen 8B (5-shot)} & \multicolumn{1}{l|}{86.2} & \multicolumn{1}{l|}{88.0} & 87.0 & \multicolumn{1}{l|}{90.6} & \multicolumn{1}{l|}{92.8} & 91.6 & \multicolumn{1}{l|}{86.5} & \multicolumn{1}{l|}{88.6} & 87.5 & \multicolumn{1}{l|}{83.3} & \multicolumn{1}{l|}{88.8} & 85.8 & \multicolumn{1}{l|}{86.0} & \multicolumn{1}{l|}{87.9} & 86.9 & \multicolumn{1}{l|}{86.7} & \multicolumn{1}{l|}{89.4} & 88.0 \\ 
\textbf{Qwen 32B (0-shot)} & \multicolumn{1}{l|}{84.8} & \multicolumn{1}{l|}{90.2} & 87.4 & \multicolumn{1}{l|}{88.8} & \multicolumn{1}{l|}{92.3} & 90.5 & \multicolumn{1}{l|}{83.8} & \multicolumn{1}{l|}{89.0} & 86.3 & \multicolumn{1}{l|}{80.8} & \multicolumn{1}{l|}{88.8} & 84.5 & \multicolumn{1}{l|}{83.3} & \multicolumn{1}{l|}{87.7} & 85.4 & \multicolumn{1}{l|}{84.6} & \multicolumn{1}{l|}{89.9} & 87.1 \\ 
\textbf{Qwen 32B (5-shot)} & \multicolumn{1}{l|}{84.6} & \multicolumn{1}{l|}{90.1} & 87.2 & \multicolumn{1}{l|}{88.2} & \multicolumn{1}{l|}{91.9} & 90.0 & \multicolumn{1}{l|}{84.0} & \multicolumn{1}{l|}{88.9} & 86.3 & \multicolumn{1}{l|}{81.1} & \multicolumn{1}{l|}{89.1} & 84.8 & \multicolumn{1}{l|}{83.3} & \multicolumn{1}{l|}{87.6} & 85.3 & \multicolumn{1}{l|}{84.5} & \multicolumn{1}{l|}{89.8} & 87.0 \\ 
\textbf{Mistral 7B (0-shot)} & \multicolumn{1}{l|}{86.0} & \multicolumn{1}{l|}{89.1} & 87.4 & \multicolumn{1}{l|}{91.0} & \multicolumn{1}{l|}{93.0} & 91.9 & \multicolumn{1}{l|}{85.6} & \multicolumn{1}{l|}{87.0} & 86.2 & \multicolumn{1}{l|}{84.9} & \multicolumn{1}{l|}{88.1} & 86.3 & \multicolumn{1}{l|}{85.6} & \multicolumn{1}{l|}{87.8} & 86.6 & \multicolumn{1}{l|}{86.8} & \multicolumn{1}{l|}{89.2} & 87.9 \\ 
\textbf{Mistral 7B (5-shot)} & \multicolumn{1}{l|}{85.9} & \multicolumn{1}{l|}{89.0} & 87.3 & \multicolumn{1}{l|}{91.1} & \multicolumn{1}{l|}{93.1} & 92.1 & \multicolumn{1}{l|}{85.6} & \multicolumn{1}{l|}{87.0} & 86.2 & \multicolumn{1}{l|}{84.7} & \multicolumn{1}{l|}{88.1} & 86.3 & \multicolumn{1}{l|}{85.4} & \multicolumn{1}{l|}{87.8} & 86.5 & \multicolumn{1}{l|}{86.8} & \multicolumn{1}{l|}{89.2} & 87.9 \\ 
\textbf{Llama 1B (Fine-tuned)} & \multicolumn{1}{l|}{93.6} & \multicolumn{1}{l|}{93.7} & 93.6 & \multicolumn{1}{l|}{97.7} & \multicolumn{1}{l|}{97.7} & 97.7 & \multicolumn{1}{l|}{92.1} & \multicolumn{1}{l|}{92.2} & 92.1 & \multicolumn{1}{l|}{95.0} & \multicolumn{1}{l|}{94.5} & 94.7 & \multicolumn{1}{l|}{93.4} & \multicolumn{1}{l|}{93.1} & 93.2 & \multicolumn{1}{l|}{94.4} & \multicolumn{1}{l|}{94.4} & 94.4 \\ 
\textbf{Qwen 0.6B (Fine-tuned)} & \multicolumn{1}{l|}{93.4} & \multicolumn{1}{l|}{93.7} & 93.5 & \multicolumn{1}{l|}{97.5} & \multicolumn{1}{l|}{97.6} & 97.5 & \multicolumn{1}{l|}{92.0} & \multicolumn{1}{l|}{92.4} & 92.1 & \multicolumn{1}{l|}{95.2} & \multicolumn{1}{l|}{95.1} & \textbf{95.1} & \multicolumn{1}{l|}{92.4} & \multicolumn{1}{l|}{92.5} & 92.4 & \multicolumn{1}{l|}{94.3} & \multicolumn{1}{l|}{94.4} & 94.3 \\ 
\textbf{MODEE-Small} & \multicolumn{1}{l|}{92.5} & \multicolumn{1}{l|}{92.6} & 92.5 & \multicolumn{1}{l|}{98.4} & \multicolumn{1}{l|}{98.3} & 98.4 & \multicolumn{1}{l|}{91.9} & \multicolumn{1}{l|}{92.6} & 92.2 & \multicolumn{1}{l|}{93.8} & \multicolumn{1}{l|}{93.7} & 93.7 & \multicolumn{1}{l|}{92.4} & \multicolumn{1}{l|}{93.8} & 93.0 & \multicolumn{1}{l|}{93.9} & \multicolumn{1}{l|}{94.2} & 94.0 \\ 
\textbf{MODEE-Base} & \multicolumn{1}{l|}{94.3} & \multicolumn{1}{l|}{94.4} & \textbf{94.3} & \multicolumn{1}{l|}{98.2} & \multicolumn{1}{l|}{98.3} & 98.2 & \multicolumn{1}{l|}{92.2} & \multicolumn{1}{l|}{92.5} & \textbf{92.3} & \multicolumn{1}{l|}{94.8} & \multicolumn{1}{l|}{94.6} & 94.6 & \multicolumn{1}{l|}{93.4} & \multicolumn{1}{l|}{93.8} & \textbf{93.6} & \multicolumn{1}{l|}{94.7} & \multicolumn{1}{l|}{94.8} & \textbf{94.7} \\ \hline
\end{tabular}
}
\caption{Performance comparison on event extraction (\%) using BERTScore. Bold indicates the best performance.}
\label{tab:my-table6}
\end{table*}

\end{document}